\documentclass{article}
\usepackage{spconf,amsmath,epsfig}
\usepackage{eufrak}
\makeatletter

\newcommand{\Rnum}[1]{\expandafter\@slowromancap\romannumeral #1@}
\usepackage{multirow}
\usepackage{color}

\begin{document}\sloppy

% Example definitions.
% --------------------
\def\x{{\mathbf x}}
\def\L{{\cal L}}

% Title.
% ------
\title{Compare more nuanced: Pairwise Alignment Bilinear Network \\ for Few-shot Fine-grained  Learning \thanks{$^*$Corresponding Author. The authors greatly appreciate the financial support from the Rail Manufacturing Cooperative Research Centre (funded jointly by participating rail organisations and the Australian Federal Government’s Business-Cooperative Research Centres Program) through Project R3.7.3 - Rail infrastructure defect detection through video analytics. }}
%\
% Single address.
% ---------------
\name{Huaxi Huang, Junjie Zhang, Jian Zhang$^{*}$, Qiang Wu, Jingsong Xu}

\address{%\textit{Global Big Data Technologies Centre, School of Electrical and Data Engineering.} \\
		{University of Technology Sydney, Australia}\\
		$\{$Huaxi.Huang@student., Junjie.Zhang@student., Jian.Zhang@, Qiang.Wu@, Jingsong.Xu@$\}$uts.edu.au
}

\maketitle

\begin{abstract}
The recognition ability of human beings is developed in a progressive way. Usually, children learn to discriminate various objects from coarse to fine-grained with limited supervision. Inspired by this learning process, we propose
a simple yet effective model for the Few-Shot Fine-Grained (FSFG) recognition, which tries to tackle the challenging fine-grained recognition task using meta-learning. The proposed method, named Pairwise Alignment Bilinear Network (PABN), is an end-to-end deep neural network. Unlike traditional deep bilinear networks for fine-grained classification, which adopt the self-bilinear pooling to capture the subtle features of images, the proposed model uses a novel pairwise bilinear pooling to compare the nuanced differences between base images and query images for learning a deep distance metric. In order to match base image features with query image features, we design feature alignment losses before the proposed pairwise bilinear pooling. Experiment results on four fine-grained classification datasets and one generic few-shot dataset demonstrate that the proposed model outperforms both the state-of-the-art few-shot fine-grained and general few-shot methods.    
\end{abstract}
\begin{keywords}
Few-shot Fine-grained , Pairwise Bilinear, Feature Alignment
\end{keywords}
\section{Introduction}
\label{sec:intro}
Fine-grained image classification aims at distinguish different sub-categories belong to the same entry-level category~\cite{WahCUB_200_2011,KhoslaYaoJayadevaprakashFeiFei_FGVC2011,KrauseStarkDengFei-Fei_3DRR2013,Horn_2015_CVPR}. This task is particularly challenging due to the low inter-category variation yet high intra-category discordance caused by the various objects posture, illumination condition and distance from the camera \textit{etc.} Compared to the part based fine-grained methods~\cite{zhang2014part,Fu_2017_CVPR}, global feature based fine-grained~\cite{Lin_2015_ICCV,Cui_2017_CVPR,Li_2018_CVPR} approaches achieve the state-of-the-art recognition performance. In addition, self-bilinear models are the most widely used approaches~\cite{Lin_2015_ICCV,Cui_2017_CVPR,Li_2018_CVPR}.  
The majority of fine-grained recognition approaches need to be fed with a large amount of training data before obtaining a decent classifier~\cite{zhang2014part,Fu_2017_CVPR,Lin_2015_ICCV,Cui_2017_CVPR,Li_2018_CVPR,Krause_2015_CVPR}. However, labelling the fine-grained data requires strong domain knowledge, \textit{e.g.}, only ornithologists can accurately identify different birds, which is significantly expensive compared to generic object recognition tasks. Moreover, in some fine-grained datasets ~\cite{zhuang2018wildfish,Horn_2018_CVPR}, the amount of the well-labelled training samples is limited, \textit{e.g.,} it is hard to collect large-scale samples of endangered species. Therefore, how to tackle the fine-grained image recognition with less training data is still an open problem.

\begin{figure}[t]
	\centerline{
		\includegraphics[width=3.4in,height=2.4in]{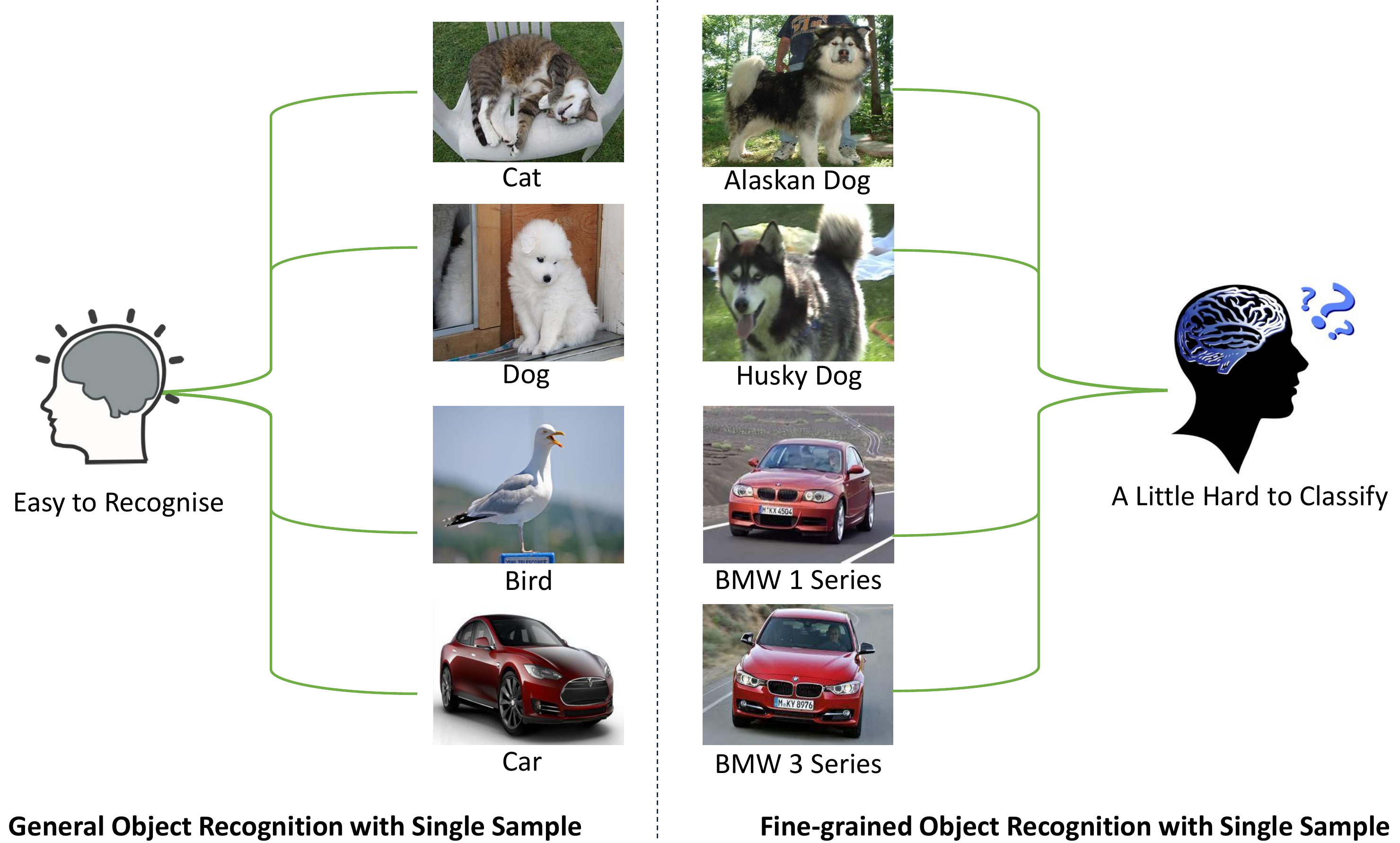}
	}
	\caption{An example of general one-shot learning (Left) and fine-grained one-shot learning (Right). For general one-shot learning, it is easy to learn the concepts of objects with only a single image. However, it is difficult to distinguish the sub-classes of specific categories with one sample.}
	\label{fig1}
\end{figure} 

\begin{figure*}[t]
	\centerline{
		\includegraphics[width=1\linewidth]{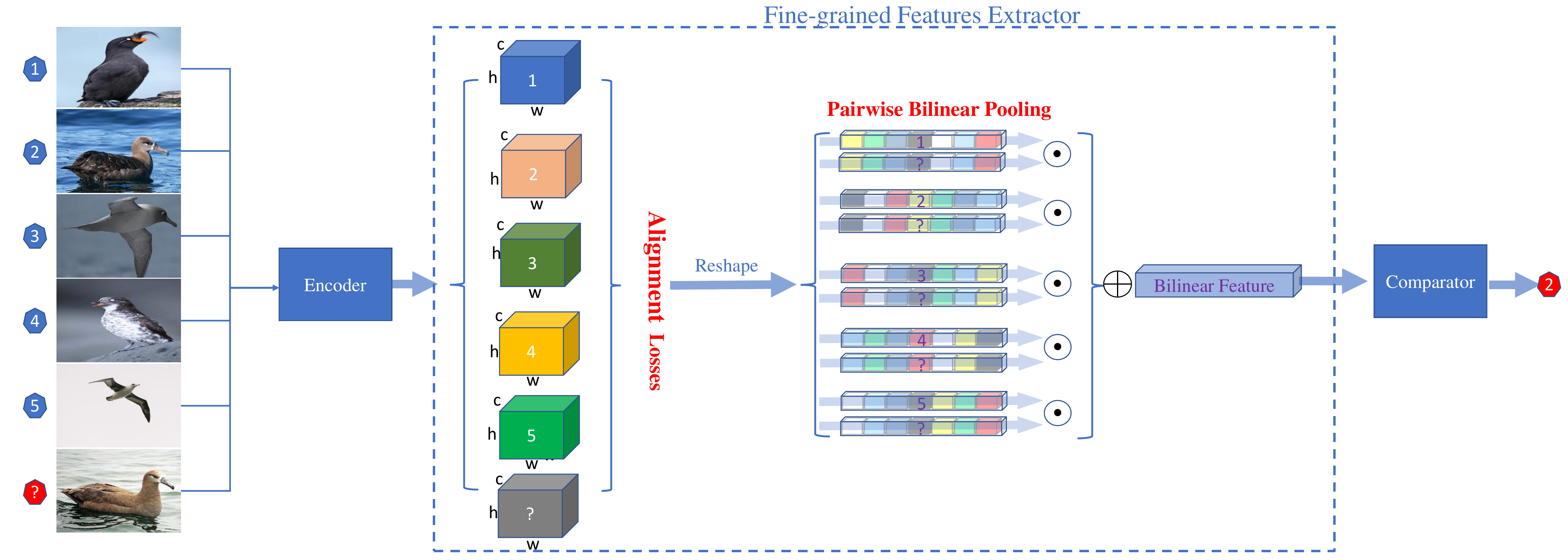}}
	\caption{The framework of PABN under the one-shot fine-grained image recognition setting. There are three parts of PABN: Encoder, Fine-grained Features Extractor, and Comparator. Encoder extracts coarse features from raw images. Fine-grained Extractor captures the subtle features further. Comparator produces the final classification results. }
	\label{fig2}
\end{figure*}

Machine few-shot learning is first proposed by Li \textit{et al.}~\cite{fei2006one} based on the Bayesian theory. Recently, due to the excellent performance of deep neural networks, machine few-shot learning~\cite{vinyals2016matching,snell2017prototypical,Sung_2018_CVPR,liu2018transductive} revives again and achieves significant improvements against previous methods. In the cognition process of human beings, preschoolers can easily distinguish the difference between `Dog' and `Horse' after seeing few samples. However, they may be confused about `Husky Dogs' and `Alaskan Dogs' with only limited samples. This can be caused by the underdeveloped ability of children to process information compared to adults, which indicates that general few-shot methods cannot cope with the fine-grained recognition task well. To this end, in this paper, we focus on dealing with the FSFG classification in a `developed' way. 

Few-Shot Fine-Grained recognition (FSFG) task is recently introduced by Wei \textit{et al.}~\cite{wei2018piecewise}. Two sub-networks are employed to jointly tackle this problem. The first is a self-bilinear encoder network, which adepts the matrix outer product operation on convolved features to capture subtle image features, while the second one is a mapping network that learns the decision boundaries of the input data. Using the meta-learning strategy on the auxiliary dataset, their model can classify different samples in the testing dataset with few labeled samples, \textit{i.e.,} few shots. 

Compared to the generic image classification, they use self-bilinear pooling to extract more informative image representations. However, for the unseen or new categories, the data distribution could be different from the training data, which means the trained self-bilinear feature extractor may fail in distinguishing these classes. 
It would be better to learn the relation or discrimination between different categories with extracting subtle features at the same time.
To solve this problem, in this paper, we propose a pairwise bilinear pooling operation between base and query images to extract fine-grained features. Meanwhile, their relations are explored by a non-linear comparator.

Most recently, Pahde \textit{et al.}~\cite{pahde2018cross} propose a cross-model FSFL method, which embeds the textual annotations and image features into a common latent space. They also introduce a discriminative text-conditional GAN for the sample generation, which selects the representative samples from the auxiliary data. However, it is both computation and time consuming to obtain rich annotations for the fine-grained samples, which we try to avoid. Yao \textit{et al.}~\cite{yao2017one} propose a one-shot fine-grained retrieval method, which employs the Convolutional and Normalization Networks. Different from their method, our model focuses on the image recognition task rather than retrieval.

There are also series of related works of meta-learning based few-shot recognition methods~\cite{vinyals2016matching,snell2017prototypical,Sung_2018_CVPR}, among which the Ration-Net~\cite{Sung_2018_CVPR} achieves the state-of-the-art performance by combining a non-linear feature encoder and a relation comparator. However, the feature extraction in Ration-Net only concatenates the base and query feature maps in the depth dimension, and cannot capture nuanced features for the fine-grained classification.  

To overcome the shortcomings of underdeveloped feature extraction in \cite{Sung_2018_CVPR} and naive self-bilinear pooling in \cite{wei2018piecewise}, we propose a novel end-to-end FSFG framework that captures the fine-grained relations among different classes. This nuanced compare ability of our models is inherently more intelligent than simply modeling the data distribution. The whole framework is shown in Figure \ref{fig2}.
More specifically, the base images and a query image are fed into the PABN simultaneously in a paired manner, followed by the encoder network to generate embedded pair features. Then a \textbf{\textit{pairwise bilinear pooling}} operation is used to extract the subtle features from these pairs. For each pair, the proposed \textit{\textbf{feature alignment losses}} are adopted to guarantee that the positions of base image features match the query ones.
Finally, pairwise bilinear features pass through the non-linear comparator, which classifies the query image into its corresponding category. In summary, the main contributions of this work are as follows:

\begin{itemize}
	\item We proposes a novel FSFG model, which mimics the advanced learning process of human beings. We propose a new pairwise bilinear pooling operation to capture the subtle differences between the base and query images.
	\item In order to acquire the accurate pairwise bilinear features, we adopt the alignment losses to regularize the embedding features.
	\item The proposed method achieves the state-of-the-art performances compared to the general few-shot learning and FSFG methods.
\end{itemize}

\section{Methodology}
\subsection{Problem Definition} \label{def}
Given a fine-grained target dataset $\mathcal{T}:$
\begin{equation}
\begin{split}
&\mathcal{T} = \left\{ \mathcal { B } = \left\{ \left(\overline{x} _ { b } , \overline{y} _ { b } \right) \right\} _ { b = 1 } ^ { K \times \tilde{C} } \right\} \cup \left\{\mathcal { N } = \left\{ \left( \overline{x} _ { v }  \right) \right\} _ { v = 1 } ^ { V } \right\}, \\
&\overline{y} _ { b } \in \{ 1 , \tilde{C} \} , \overline { x } \in   \mathcal { R } ^ { N }, \mathcal{B} \cap \mathcal{N} = \emptyset, V \gg K \times \tilde{C}.
\end{split}
\end{equation} 
For the FSFG task, the target dataset $\mathcal{T}$ contains two parts: the labeled subset $\mathcal{ B }$ and the unlabeled subset $\mathcal{ N }$. The model needs to classify the unlabeled data from $\mathcal{ N }$ ($\overline{x} _ { v }$ represents the raw image) according to the few labeled data from $\mathcal{ B }$ (where $\overline{x} _ { b }$ denotes the image and $\overline{y} _ { b }$ is the label of this image). If the labeled data in the target dataset contains $K$ labeled images for each of $\tilde{C}$ different categories, this problem called $\tilde{C}$-way-$K$-shot problem. 

In order to get a perfect model that can identify the unlabeled images from the target dataset $\mathcal{ N }$. Few-shot learning usually employs a fully annotated dataset which has similar property or data distribution with $\mathcal{ T }$ as the auxiliary dataset $\mathcal{A}$: 
\begin{equation}
\begin{split}
&\mathcal{A} = \left\{ \mathcal { S } = \left\{ \left( x _ { i } , y _ { i } \right) \right\} _ { i = 1 } ^ { I } \right\} \cup \left\{\mathcal { Q } = \left\{ \left( x _ { j } , y _ { j } \right) \right\} _ { j = 1 } ^ { J } \right\}, \\
& y_{i},{y}_{j} \in \{ 1 , {C} \} , x \in   \mathcal { R } ^ { N }, \mathcal{S} \cap \mathcal{Q} = \emptyset, \mathcal{A} \cap \mathcal{T} = \emptyset.
\end{split}
\end{equation}
Where  $x_{i}$ and ${x}_{j}$ represent images, $y_{i}$ and ${y}_{j}$ represent image labels. In each round of training, the auxiliary dataset $\mathcal{A}$ is randomly separated into two parts: Support dataset $\mathcal{ S }$ and Query dataset $\mathcal{ Q }$. With setting $I = K \times \tilde{C}$, we can simulate the composition of target dataset in each iteration. Then $\mathcal{ A}$ is used to learn a meta-learner $\mathfrak{F}$ which can transfer the knowledge from $\mathcal{A}$ to target data $\mathcal{T}$. Once the meta-learner is trained, it could be fine-tuned using labeled target dataset $\mathcal{ B }$. Finally, the meta-learner could classify the samples from the unlabeled data $\mathcal{ N }$ into their corresponding categories.
This training setting that mimics the few-shot setting of target problem is widely used in meta-learner training~\cite{vinyals2016matching,Sung_2018_CVPR,wei2018piecewise}.

\subsection{Framework}
The whole framework of PABN is shown in Figure~\ref{fig2}. Different from traditional few-shot embedding structures~\cite{vinyals2016matching,snell2017prototypical,Sung_2018_CVPR}, 
we add the fine-grained image feature extractors as shown in the dotted line box which is our main contribution. In addition, we modify the non-linear comparator~\cite{Sung_2018_CVPR} and apply it to our fine-grained task. 
Fine-grained features extractor can be divided into two structures: alignment loss regularization and pair-wise bilinear pooling layer. The former aims to match the features of the same position in the embedded images features. For example, the features of the bird's head in the target dataset $\mathcal{B}$ should match the query bird's head features from $\mathcal{ Q }$. The latter pairwise bilinear pooling layer is designed to extract the second-order comparative features from pairs of base images (like samples from $\mathcal{ B }$) and query images (like samples from $\mathcal{ N }$).

Pairwise bilinear pooling layer is the core component of PABN model which captures the nuanced comparative features of image pairs and therefore decides the relations between base and query images which is crucial to the classifier. However, if the pair of the images are not well matched, this pairwise bilinear pooled features cannot result in the maximum classification performance gain. Thus we propose two feature alignment losses to guarantee the registration between pairs of images.
In next section, we will firstly introduce the pairwise bilinear pooling layer, then we will present the feature alignment regularization with two alignment losses.
\subsection{Pairwise Bilinear Pooling Layer}
Original Bilinear CNN images recognition can be defined as a quadruple:
\begin{equation}
\begin{split}
&\textit{B-CNNs} = (\mathfrak{E}_{\Rnum{1}}, \mathfrak{E}_{\Rnum{2}}, \mathfrak{f}_{b}, \mathcal{C}),  \\
&\mathfrak{E}: \mathcal{I} \longrightarrow \mathcal{X} \in \mathcal{R}^{c \times h\times w}, \\
&\mathfrak{f}_{b}(\mathcal{I},\mathfrak{E}_{\Rnum{1}}, \mathfrak{E}_{\Rnum{2}}) = \frac{1}{hw} \sum_{i=1}^{hw} f_{\alpha,i} f_{\beta,i}^{T}.  
\label{eq3}
\end{split} 
\end{equation}
$\mathfrak{E}_{\Rnum{1}}$ and $\mathfrak{E}_{\Rnum{2}}$ are two encoders. $\mathfrak{f}_{b}$ is the self-bilinear pooling and $\mathcal{C}$ represents a classifier. $\mathcal{I} \in \mathcal{R}^{H \times W \times C}$ is a image that has $H$ height, $W$ width and $C$ color channels. Through encoder $\mathfrak{E}$, the input image is transformed into a tensor $\mathcal{M} \in \mathcal{R}^{h \times w \times c}$ which has $c$ feature channels and $h,w$ indicate the hight and width of the embedded feature map. Given two specific functions $\mathfrak{E}_{\Rnum{1}}: \mathcal{S} \longrightarrow \mathcal{X}_{\alpha} \in \mathcal{R}^{c_1 \times h\times w} $ and $\mathfrak{E}_{\Rnum{2}}: \mathcal{S} \longrightarrow \mathcal{X}_{\beta} \in \mathcal{R}^{c_2 \times h \times w} $.
$f_{\alpha,i} \in \mathcal{R}^{c_1 \times 1}$  and  $f_{\beta,i} \in \mathcal{R}^{c_2 \times 1}$ denote feature vectors at specific location in each feature matrix $\mathcal{X}_{\alpha}$ and $\mathcal{X}_{\beta}$  with $i \in [1,hw]$. The pooled feature is a $c_1 \times c_2$ vector. $\mathcal{C}$ is a fully-connected layer with the cross-entropy training loss between self-bilinear feature and image label.

The self-bilinear operates on pairs of embedded features from the same image. However, in our\textbf{\textit{ pairwise bilinear pooling}}, given a pair of image $\mathcal{ I_{A} }$ (e.g., $\mathcal{ I_{A} } \in \mathcal{S}$ ) and image $\mathcal{I_{B} }$ (e.g., $\mathcal{ I_{B} } \in \mathcal{Q}$ ), an encoder $\mathfrak{\tilde{E}}$, pairwise bilinear pooling $\mathfrak{f}_{pb}$ can be defined as
\begin{equation}
\begin{split}
&\mathfrak{f}_{pb}(\mathcal{I_{A}},\mathcal{I_{B}}, \mathfrak{\tilde{E}}) = \mathfrak{\tilde{E}}(\mathcal{ I_{A} })\mathfrak{\tilde{E}}(\mathcal{ I_{B} })^{T},\\
&\mathfrak{\tilde{E}}: \mathcal{I} \longrightarrow \mathcal{X} \in \mathcal{R}^{c \times hw}.
\label{eq4}
\end{split} 
\end{equation}
After obtaining this pairwise bilinear vectors, a sigmoid activation is used to generate the relation scores of the compared pairs. The relation scores are then passed to the final comparator. 

Note that in our pairwise bilinear pooling, we only have one shared embedding functions $\mathfrak{\tilde{E}}$. Different from the self-bilinear pooling that operates on the same input image, pairwise bilinear pooling uses matrix outer product on two disparate samples. 
The training loss in our bilinear comparator is mean square error (MSE) loss which regresses the relation score to the images label similarity as discussed in~\cite{Sung_2018_CVPR}.
In this way, we can capture the fine-grained second-order comparative features in a pair-wise manner.
\subsection{Feature Alignment Loss}
In Equation~\ref{eq3}, self-bilinear pooling operates on the same image which means in any location of the embedded features map, the operates features should be aligned. However, our proposed pairwise bilinear pooling conducts on different samples, thus the encoded features may not always matched. In order to overcome this problem, we design two feature alignment losses as follows:
\begin{equation}
\begin{split}
&\textit{Align}_{loss_{1}}(\mathcal{I_{A}},\mathcal{I_{B}}, \mathfrak{\tilde{E}}) = MSE( \mathfrak{\tilde{E}}(\mathcal{ I_{A}}),\mathfrak{\tilde{E}}(\mathcal{ I_{B} })). 
\label{eq5}
\end{split} 
\end{equation}
the first $\textit{Align}_{loss_{1}}$ loss is a rough approximation of two embedded image descriptors which minimzing the Euclidean distances of all elements of two features. 
\begin{equation}
\begin{split}
%&\textit{Align}_{loss_{1}}(\mathcal{I_{A}},\mathcal{I_{B}}, \mathfrak{\tilde{E}}) = MSE( \mathfrak{\tilde{E}}(\mathcal{ I_{A}}),\mathfrak{\tilde{E}}(\mathcal{ I_{B} }));\\
&\textit{Align}_{loss_{2}}(\mathcal{I_{A}},\mathcal{I_{B}}, \mathfrak{S}) = MSE( \mathfrak{S}(\mathcal{ I_{A}}),\mathfrak{S}(\mathcal{ I_{B} })),\\
&MSE( \mathfrak{S}(\mathcal{ I_{A}}),\mathfrak{S}(\mathcal{ I_{B} })) = \sum_{1}^{hw}{(\mathfrak{S}(\mathcal{ I_{A} })-\mathfrak{S}(\mathcal{ I_{B} }))^{2}},\\
&\mathfrak{{S}}(\mathcal{I})= \sum_{1}^{c}{\mathfrak{\tilde{E}}(\mathcal{ I })}, \mathfrak{\tilde{E}}: \mathcal{I} \longrightarrow \mathcal{X} \in \mathcal{R}^{c \times hw}.
%&MSE( \mathfrak{\tilde{E}}(\mathcal{ I_{A}}),\mathfrak{\tilde{E}}(\mathcal{ I_{B} })) = {(\mathfrak{\tilde{E}}(\mathcal{ I_{A} })-\mathfrak{\tilde{E}}(\mathcal{ I_{B} }))^{2}}. 
\label{eq6}
\end{split} 
\end{equation}
the second ${Align_{loss_{2}}}$ loss is a more concise feature alignment loss
, where we sum all the raw features along the third-channel first and then measures the MSE of summed features as Equation~\ref{eq6} indicates.

By training with the proposed alignment losses, we encourage the network to automatically learn the matching features to generate a better pairwise bilinear feature.
\section{Experiment}
In this section, we evaluate the proposed PABN on four widely used fine-grained datasets and one generic few-shot dataset. First, we will give a brief introduction to these datasets. Then we introduce the experiment setup in detail. Finally, we analyze the experimental results of the proposed models and compare with other few-shot learning approaches.
\subsection{Dataset}
In our experiments, we utilize five datsets to investigate the proposed models:
\begin{itemize}
	\item CUB Birds~\cite{WahCUB_200_2011} contains 200 categories of birds and totally 11,788 images.
	\item DOGS~\cite{KhoslaYaoJayadevaprakashFeiFei_FGVC2011} contains 120 categories of dogs and totally 20,580 images.
	\item CARS~\cite{KrauseStarkDengFei-Fei_3DRR2013} contains 196 categories of cars and totally 16,185 images.
	\item NABirds~\cite{Horn_2015_CVPR} contains 555 categories of north American birds and totally 48,562 images.
	\item MiniImageNet~\cite{vinyals2016matching} consists 100 categories of 60,000 images. Each class has 600 examples. 	
\end{itemize}
\begin{table}[t]
	\begin{center}
		\fontsize{9.5}{14}\selectfont
		\caption{The class split for four fine-grained datasets. $C_{total}$ is the original number of categories in the datasets, $C_{\mathcal{ A}}$ is the number of categories in separated auxiliary datasets and $C_{\mathcal{ T }}$ is the number of categories in target datasets.} \label{tab:cap}
	      
		\begin{tabular}{ | c | c | c | c | c| } \hline
			\text { Dataset } & { \text{CUB Birds}  } & { \text { DOGS } } & { \text { CARS } } & {\text{NABirds}} \\ 
			\hline \hline
			$C _ { \text { total } }$ & { 200 } & { 120 } & { 196 } &{555} \\ 
			$C _ { \mathcal{ A}  }$ & { 150 } & { 90 } & { 147 }&{416} \\ 
			$C _ { \mathcal { T } }$ & { 50 } & { 30 } & { 49 }&{139} \\ \hline 
		\end{tabular} 
	\end{center}
\end{table}
In Section~\ref{def}, we randomly divide these datasets into two disjoint sub-datasets: the auxiliary dataset $\mathcal{ A}$ and the target dataset $\mathcal{ T }$ as shown in Table~\ref{tab:cap}. For CUB Birds, DOGS and CARS datasets, we follow Wei's~\cite{wei2018piecewise} separation. 
For MiniImageNet, we followed the separation of \cite{Sung_2018_CVPR} which adopts 64, 16, and 20 classes as training set, validation set and testing set, respectively. Notice that the validation set is only used for monitoring the generalisation of performance.

\subsection{Experimental Setup}

\begin{table*}[t]  
	\centering  
	\fontsize{8.8}{16.5}\selectfont  
	\caption{Few-shot classification accuracy (\%) comparison on four fine-grained datasets.  The highest-accuracy methods are highlighted. The second highest-accuracy methods are labeled with the underline. `-' denotes not reported. All results are with $95\%$ confidence intervals where reported.} \label{tab2}
	\label{BCN}  
	\begin{tabular}{|c|c|c|c|c|c|c|c|c|}  
		\hline
		\multirow{2}{*}{Methods} & 	\multicolumn{2}{|c|}{CUB Birds}&\multicolumn{2}{|c|}{CARS }&\multicolumn{2}{|c|}{DOGS}&\multicolumn{2}{|c|}{NABirds}  \cr\cline{2-9}
		&1-shot & 5-shot & 1-shot & 5-shot&1-shot & 5-shot & 1-shot & 5-shot \\
		\hline  \hline
		PCM~\cite{wei2018piecewise} & 42.10$\pm$1.96 & 62.48$\pm$1.21& 29.63$\pm$2.38 & 52.28$\pm$1.46 & 28.78$\pm$2.33 & 46.92$\pm$2.00  & - & -\\ \hline
		Ration-Net &63.77$\pm$1.37 &74.92$\pm$0.69& \underline{56.28$\pm$0.45} & 68.39$\pm$0.21 & 51.95$\pm$0.46 & 64.91$\pm$0.24 & 65.17$\pm$0.47 & 78.35$\pm$0.21 \\ \hline
		PABN$_{w/o}$ &\underline{65.99$\pm$1.35 } &\textbf{76.90$\pm$0.21 } &55.65$\pm$0.42   &67.29$\pm$0.23   &\underline{54.77$\pm$0.44 } &65.92$\pm$0.23  &\textbf{67.23$\pm$0.42 } &\underline{79.25$\pm$0.20 }   \\ \hline
		PABN$_{loss1}$ &65.04$\pm$0.44  & 76.46$\pm$0.22  &55.89$\pm$0.42   &\underline{68.53$\pm$0.23 } &54.06$\pm$0.45  &\underline{65.93$\pm$0.24 } &66.62$\pm$0.44  &\textbf{79.31$\pm$0.22 }  \\ \hline  
		PABN$_{loss2}$ &\textbf{66.71$\pm$0.43 } & \underline{76.81$\pm$0.21 } &\textbf{56.80$\pm$0.45 } &$\textbf{68.78$\pm$0.22 }$ &\textbf{55.47$\pm$0.46 } &\textbf{66.65$\pm$0.23 }&\underline{67.02$\pm$0.43 } &79.02$\pm$0.21   \\ \hline
	\end{tabular}  
\end{table*}

In each round of training and testing, for one-shot image recognition, the base sample number in each class equals 1 (in both $\mathcal{ B }$ and $\mathcal{ S }$, $K = 1$). Therefore we use the embedded features of these base sample as the classes' features ($\mathfrak{\tilde{E}}(\mathcal{ I_{A} })$). For few-shot image recognition, we extract the classes' features by summing all the embedded features in each category.
We compared four variations of the proposed PABN models: PABN$_{w/o}$, PABN$_{loss1}$ and PABN$_{loss2}$. PABN$_{w/o}$ represents the model that does not use alignment loss on embedded pair features. PABN$_{loss1}$, and PABN$_{loss2}$ are the models which adopt the alignment loss $Align_{loss_{1}}$ and $Align_{loss_{2}}$ separately in alignment layer. After pairwise bilinear pooling, we conduct normalization operation on pairwise bilinear features as \cite{Lin_2015_ICCV} did.

In all our PABN models and Rational Network, we conduct 5-way-1-shot and 5-way-5-shot settings. Both of 5-way-1-shot and 5-way-5-shot experiments have 15 query images which means there are $15 \times 5 + 1 \times 5 = 80$ images and $15 \times 5 + 5 \times 5 = 100$ images separately for 5-way-1-shot and 5-way-5-shot in each mini-batches. We resize all the input images from all datasets to $84 \times 84$. All experiments use Adam optimize method with initial learning rate 0.001 and all models are trained end-to-end from scratch. We initialize all networks randomly without involving additional datasets.
\begin{table}[t]  
	\centering  
	\fontsize{9}{14}\selectfont  
	\label{BCNN}  
	\caption{Experiments results on MiniImageNet dataset. The highest-accuracy methods are highlighted and the second highest-accuracy methods are labeled with the underline.  With $95\%$ confidence intervals.} \label{tab3}
	\begin{tabular}{|c|c|c|}  
		\hline
		\multirow{2}{*}{Methods} & 	\multicolumn{2}{|c|}{MiniImageNet 5-way}  \cr \cline{2-3}
		&1-shot & 5-shot \\
		\hline  \hline
		%ProtoNetwork~\cite{snell2017prototypical} & 49.42 $\pm$ 0.78\% & \textbf{68.20 $\pm$ 0.66\%} \\ \hline
		Ration-Net~\cite{Sung_2018_CVPR} & 50.44 $\pm$ 0.82\% & \underline{65.32 $\pm$ 0.70\%} \\ \hline
		PABN$_{w/o}$ & \textbf{51.87 $\pm$ 0.45\% } & 64.95 $\pm$ 0.71\%   \\ \hline
		PABN$_{loss1}$ & 50.55 $\pm$ 0.44\%  & 64.80 $\pm$ 0.75\%     \\ \hline  
		PABN$_{loss2}$ & \underline{50.94 $\pm$ 0.43\% }  &  \textbf{65.37 $\pm$ 0.68\%  }\\ \hline
	\end{tabular}  
\end{table} 
\subsection{Results and Analysis}\label{Res}
To the best of our knowledge, there are few methods proposed for Few-shot Fine-grained image recognition~\cite{wei2018piecewise,pahde2018cross,yao2017one}. \cite{pahde2018cross} uses larger auxiliary dataset than our methods and \cite{yao2017one} is only applied for image retrieval task. It is unfair to compare with these methods directly. Therefore we compare our PABN with Piecewise Classifier Mapping (PCM) \cite{wei2018piecewise} which is the first FSFG method. Moreover, we also compare our methods with the state-of-the-art generic few-shot learning method Ration-Net~\cite{Sung_2018_CVPR}. Original Rational Network does not report the results on four fine-grained datasets under the few-shot setting. We use the open source code of Rational Network to conduct the FSFG image recognition on these datasets.

We show the experimental results of five compared models in Table 2. As we can see, the proposed PABN models achieve siginificant improvements on both 1-shot and 5-shot recognition tasks on four fine-grained datasets compared to the state-of-the-art FSFG method and the state-of-the-art generic few-shot method which indicates the effectiveness of proposed framework. In addition, PABN models and Ration-Net obtain around 10 to 20 percent higher in recognition accuracy than PCM which demonstrates that a leaned non-linear comparator outperforms a plain linear classifier.

%PABN models outperform Ration-Net under various experimental settings. 
Specifically, without feature alignment, PABN$_{w/o}$ achieves higher averaged accuracies than Ration-Net on CUB Birds, CARS and DOGS, except on CARS data that is nearly $0.1\%$ lower in accuracy than Ration-Net. Nevertheless, by adding alignment layer with two alignment losses, PABN$_{loss2}$ and PABN$_{loss1}$ obtain higher classify accuracies for 1-shot and 5-shot on CARS separately which indicates that well-matched pairwise bilinear features can produce better recognition performance for FSFG tasks.
It can be observed that PABN$_{loss2}$ achieves the best or second best classification performance on almost datasets compared to PABN$_{loss1}$ under different experimental settings. This indicates that a more precise feature alignment can result in a better performance of pairwise bilinear pooling.

For a further analysis of our models, we conduct an additional experiment on MiniImageNet~\cite{vinyals2016matching} dataset which is a standard generic few-shot learning dataset. Form Table~\ref{tab3}, it can be observed that our PABN models achieve higher performance than Rational Network. In detail, in 1-shot recognition scene, all the PABN models outperform Rational Network with higher accuracy and lower standard deviations. As for the 5-shot setting, PABN$_{loss2}$ achieves the state-of-the-art performance where other models are slightly lower in accuracies than Rational Network. Moreover, PABN$_{loss2}$ achieves the best classification performance in 5-shot learning and second best classification performance in 1-shot learning.
That also demonstrates that a concise matching of compared features can further improve the performance.

\section{Conclusion}
In this paper, we propose a novel few-shot fine-grained image recognition method which is inspired by the advanced information processing ability of human beings. The main contribution is the pairwise bilinear pooling, which extracts the second-order comparative features for the pair of base images and query images. Moreove, in order to get a more precise comparative feature, we propose two feature alignment losses to match the embedded base image features with query image features. Through comprehensive experiments on five widely used datasets, we verify the effectiveness of the proposed method.
In our future work, we would like to design a more accurate feature matching model as Section~\ref{Res} discussed. %In addition, we would like to learn a more informative base data representation for few-shot or zero-shot setting as in \cite{garcia2018few,liu2018transductive}.
\bibliographystyle{IEEEbib}
\bibliography{icme2019template}

\end{document}